\documentclass[11pt,a4paper]{article}
\usepackage[hyperref]{emnlp2020}
\usepackage{times}
\usepackage{latexsym}

\usepackage{microtype}

\aclfinalcopy 
\def\aclpaperid{26} 

\usepackage{tikz}
\usepackage{pgfplots}
\pgfplotsset{compat=1.13}
\usepackage{multirow}
\usepackage{multicol}
\usepackage{amsmath}
\usepackage{amssymb}
\usepackage{subcaption}
\usepackage{arydshln}

\title{Self-Paced Learning for Neural Machine Translation}

\author{Yu Wan$^{\dagger\ddagger}$~~~Baosong Yang$^\ddagger$\thanks{~~Baosong Yang and Derek F. Wong are co-corresponding authors. Work was done when Yu Wan was interning at DAMO Academy, Alibaba Group.}~~~Derek F. Wong$^{\dagger *}$~~~Yikai Zhou$^\dagger$\\~~~\textbf{Lidia S. Chao}$^\dagger$~~~\textbf{Haibo Zhang}$^\ddagger$~~~\textbf{Boxing Chen}$^\ddagger$\\
  $^\dagger$NLP$^2$CT Lab,
  University of Macau\\
  {\tt nlp2ct.\{ywan,yzhou\}@gmail.com, \{derekfw,lidiasc\}@umac.mo} \\
  $^\ddagger$Alibaba Group\\
  {\tt \{yangbaosong.ybs,zhanhui.zhb,boxing.cbx\}@alibaba-inc.com}}

\date{}

\begin{document}
\maketitle
\begin{abstract}
Recent studies have proven that the training of neural machine translation (NMT) can be facilitated by mimicking the learning process of humans.
Nevertheless, achievements of such kind of curriculum learning rely on the quality of artificial schedule drawn up with the hand-crafted features, e.g. sentence length or word rarity.
We ameliorate this procedure with a more flexible manner by proposing self-paced learning, where NMT model is allowed to 1) automatically quantify the learning confidence over training examples; and 2) flexibly govern its learning via regulating the loss in each iteration step. 
Experimental results over multiple translation tasks demonstrate that the proposed model yields better performance than strong baselines and those models trained with human-designed curricula on both translation quality and convergence speed.\footnote{Our codes:  \href{https://github.com/NLP2CT/SPL_for_NMT}{https://github.com/NLP2CT/SPL\_for\_NMT}.}

\end{abstract}

\section{Introduction}
Neural machine translation (NMT) has achieved promising results with the use of various optimization tricks~\cite{hassan2018achieving,Chen:2018:ACL,xu2019leveraging,li2019neuron,Yang2020improve}. 
In spite of that, these techniques lead to increased training time and massive hyper-parameters, making the development of a well-performed system expensive~\cite{popel2018training,ott2018scaling}.

As an alternative mitigation, curriculum learning~\citep[CL,][]{elman1993learning,bengio2009curriculum} has shown its effectiveness on speeding up the convergence and stabilizing the NMT model training~\cite{zhang2018empirical,platanios2019competence}. 
CL teaches NMT model from easy examples to complex ones rather than equally considering all samples, where the keys lie in the definition of ``difficulty'' and the strategy of curricula design~\cite{krueger2009flexible,kocmi2017curriculum}. Existing studies artificially determine data difficulty according to prior linguistic knowledge such as sentence length (SL) and word rarity (WR)~\cite{platanios2019competence,zhang2019curriculum,zhou2020uncertainty}, and manually tune the learning schedule~\cite{liu2020norm,fomicheva2020unsupervised}. 
However, neither there exists a clear distinction between easy and hard examples~\cite{kumar2010self}, nor these human intuitions exactly conform to effective model training~\cite{zhang2018empirical}.

Instead, we resolve this problem by introducing self-paced learning~\cite{kumar2010self}, where the emphasis of learning can be dynamically determined by model itself rather than human intuitions.
Specifically, our model measures the level of confidence on each training example~\cite{gal2016dropout,xiao2019quantifying}, where an easy sample is actually the one of high confidence by the current trained model.
Then, the confidence score is served as a factor to weight the loss of its corresponding example.
In this way, the training process can be dynamically guided by model itself, refraining from human predefined patterns. 

We evaluate our proposed method on IWSLT15 En$\Rightarrow$Vi, WMT14 En$\Rightarrow$De, as well as WMT17 Zh$\Rightarrow$En translation tasks.
Experimental results reveal that our approach consistently yields better translation quality and faster convergence speed than \textsc{Transformer}~\cite{vaswani2017attention} baseline and recent models that exploit CL~\cite{platanios2019competence}.
Quantitative analyses further confirm that the intuitive curriculum schedule for a human does not fully cope with that for model learning.

\section{Self-Paced Learning for NMT}

\begin{figure}[t]
    \centering
    \includegraphics[width=1.0\columnwidth]{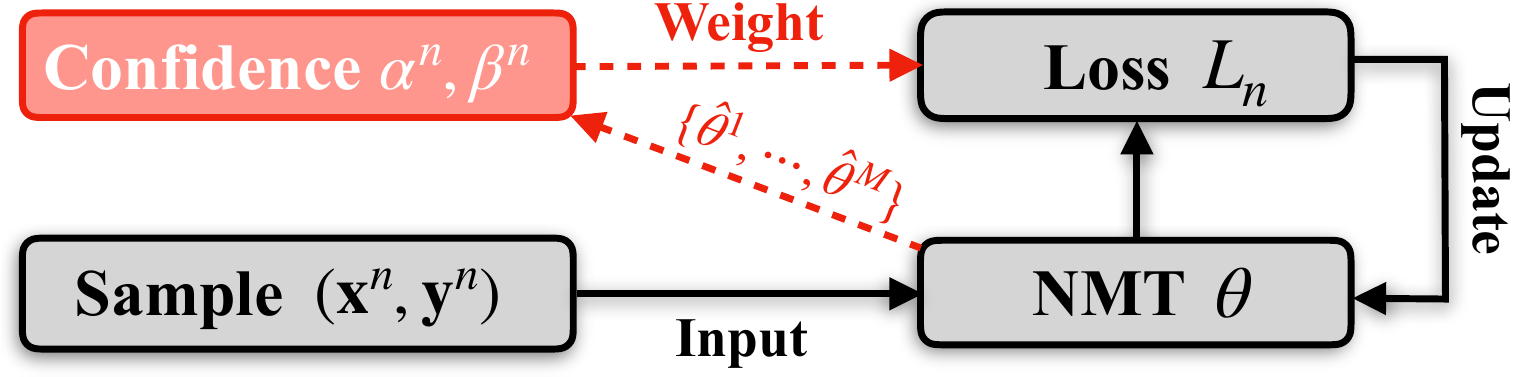}
    \caption{Illustration of the  proposed self-paced learning. The black components compose the vanilla NMT training process, while our model (red) assigns confidence scores for each input to weight its loss.}
    \label{fig:model}
\end{figure}

As mentioned above, translation difficulty for humans may not match that for neural networks.
Even if these artificial supervisions are feasible, the long sequences or rare tokens are not always ``difficult'' as the model competence increases.
From this view, we design a self-paced learning algorithm that offers NMT the abilities to 1) estimate the confidences over samples appropriated for the current training state; and 2) automatically control the focus of learning through regulating the training loss, as illustrated in Fig.~\ref{fig:model}.

\subsection{Confidence Estimation}
\label{sec.con}

We propose to determine the learning emphasis according to the model confidence~\cite{ueffing2005word,soricut2010trustrank},  which quantifies whether the current model is confident or hesitant on translating the training samples.
The model confidence can be quantified by Bayesian neural networks~\cite{buntine1991bayesian,neal1996bayesian}, which place distributions over the weights of network.
For efficiency, we adopt widely used Monte Carlo dropout sampling~\cite{gal2016dropout} to approximate Bayesian inference. 

Given current NMT model parameterized by $\mathbf{\theta}$ and a mini-batch consisting of $N$ sentence pairs $\{(\mathbf{x}^1,\mathbf{y}^1),\cdots,(\mathbf{x}^N,\mathbf{y}^N)\}$, we first perform $M$ passes through the network, where the $m$-th pass $\hat{\mathbf{\theta}}^m$ randomly deactivates part of neurons. Thus, each example yields $M$ sets of conditional probabilities. The lower variance of translation probabilities reflects higher confidence that the model has with respect to the instance~\cite{dong2018confidence,wang2019improving}. 
We propose multi-granularity strategies for confidence estimation:

\paragraph{Sentence-Level Confidence (SLC)} A natural choice for measuring the confidence of sentence pair $(\mathbf{x}^n,\mathbf{y}^n)$ is to assess the variance of translation probability $\text{Var}\{P(\mathbf{y}^n|\mathbf{x}^n,\hat{\mathbf{\theta}}^m)\}_{m=1}^M$. Accordingly, confidence score $\hat{\alpha}^n$ can be formally expressed as: 
\begin{align}
    \hat{\alpha}^n & = (1-\text{Var}\{P(\mathbf{y}^n|\mathbf{x}^n,\hat{\mathbf{\theta}}^m)\}_{m=1}^M)^k,
    \label{eq.slc}
\end{align}
Here, we assign a hyper-parameter  $k$  to scale the gap between scores of confident and unconfident examples.
The larger absolute value of $k$ represents higher discriminative manner and vice versa.
In some extreme cases, all the confidence scores in a mini-batch may tend to small or big value, e.g. the estimation at the early stage of the training.\footnote{When implementing the computation of SLC\&TLC scores, we use negative log-likelihood values instead of conventional probabilities. Besides, we use the maximum value to refactorize them within $[0, 1]$ by division.}
In order to stabilize the training process and maintain the same loss scale as conventional model, we normalize the confidence scores by $softmax$:
\begin{align}
    \alpha^n & = \frac{\exp(\hat{\alpha}^n)}{\sum_{t=1}^N \exp(\hat{\alpha}^t)}.
    \label{eq.slc_norm}
\end{align}

\paragraph{Token-Level Confidence (TLC)} Intuitively, confidence scores can be evaluated at more fine-grained level.
We extend our model into token-level so as to estimate the confidence on translating each element in target sentence $\mathbf{y}^n$. The confidence $\hat{\beta}^n_j$ of the $j$-th token $\mathbf{y}^n_j$ is:
\begin{align}
    \hat{\beta}^n_j & = (1-\text{Var}\{P(\mathbf{y}^n_j|\mathbf{x}^n,\mathbf{y}^n_{<j},\hat{\mathbf{\theta}}^m)\}_{m=1}^M)^k,
    \label{eq.tlc}
\end{align}
where $\text{Var}\{P(\mathbf{y}^n_j|\mathbf{x}^n,\mathbf{y}^n_{<j},\hat{\mathbf{\theta}}^m)$ denotes the variance of the translation probability with respect to $\mathbf{y}^n_j$.
Similar to sentence-level strategy, the confidence scores of tokens are normalized as:
\begin{align}
    \beta^n_j & = \frac{\exp(\hat{\beta}^n_j)}{\sum_{t=1}^J \exp(\hat{\beta}^n_t)},
    \label{eq.tlc_norm}
\end{align}
where $J$ indicates the length of target sentence $\mathbf{y}^n$.

\subsection{Training Strategy}
A larger confidence score indicates that the current model is confident on the corresponding example.
Therefore, the model should learn more from the  predicted loss.
In order to govern the learning schedule automatically, we leverage the confidence scores as factors to weight the loss, thus controlling the update at each time step.
To this end, the sentence log-likelihood can be defined as:
\begin{align}
    \mathcal{L}_n & = \sum_{j=1}^{J}\beta^n_j\log P(\mathbf{y}^n_j|\mathbf{x}^n,\mathbf{y}^n_{<j},\mathbf{\theta}), 
    \label{eq.loss_sent}
\end{align}
Finally, the loss of a batch is calculated as:
\begin{align}
    \mathcal{L} & = - \sum_{n=1}^{N}\alpha^n\mathcal{L}_n.
    \label{eq.loss_batch}
\end{align}
At the early stage of the study, the model learns more from confident samples, thus accelerating the training.
The hesitant samples are not completely ignorant, but relatively few can be learned.
As training proceeds, the loss of high-confidence samples gradually reduce, and the model will pay more attention on ``complex'' samples with low prediction accuracy, thus raising their confidence.
In this way, the loss of different samples are dynamically revised, eventually balancing the learning. 

Contrast to related studies~\cite{zhang2018empirical,zhang2019curriculum,kumar2019reinforcement,platanios2019competence} which adopt CL into NMT with predefined patterns, the superiority of our model lies in its flexibility on both learning emphasis and strategy.
Several researchers may concern about the processing speed when integrating Monte Carlo Dropout sampling.
Contrary to prior studies which estimate confidence during inference~\cite{dong2018confidence,wang2019improving}, we only perform forward propagation $M=5$ times in training time, which avoids the auto-regressive decoding for efficiency.

\section{Experiments}

We evaluate our method upon \textsc{Transformer}-\emph{Base}/\emph{Big} model~\cite{vaswani2017attention} and conduct experiments on IWSLT15 English-to-Vietnamese (En$\Rightarrow$Vi), WMT14 English-to-German (En$\Rightarrow$De) and WMT17 Chinese-to-English (Zh$\Rightarrow$En) tasks. 
For fair comparison, we use the same experimental setting as~\newcite{platanios2019competence} for En$\Rightarrow$Vi and follow the common configuration in~\newcite{vaswani2017attention} for En$\Rightarrow$De and Zh$\Rightarrow$En.

During training, we apply 0.3 dropout ratio and batch size as 4,096 for En$\Rightarrow$Vi task, and experiments are conducted upon one Nvidia GTX1080Ti GPU device.
For En$\Rightarrow$De and Zh$\Rightarrow$En task, we use 32,768 as batch size, and use four Nvidia V100 GPU devices for experiments.
We use beam size as 4, 5, 10, and decoding alpha as 1.5, 0.6, 1.35 for each task, respectively~\cite{vaswani2017attention}.
We compare our models with two baselines:
\begin{itemize}
    \item \emph{Base} and \emph{Big} represent the vanilla \textsc{Transformer}~\cite{vaswani2017attention} models. 
    \item \textsc{+CL} is the recent NMT model that exploits CL~\cite{platanios2019competence}. Difficulty of each training sample is estimated according to its sentence length (SL) or averaged word rarity (WR). The curriculum schedule depends on the number of training step.
\end{itemize}

\subsection{Confidence/Unconfidence Balancing}

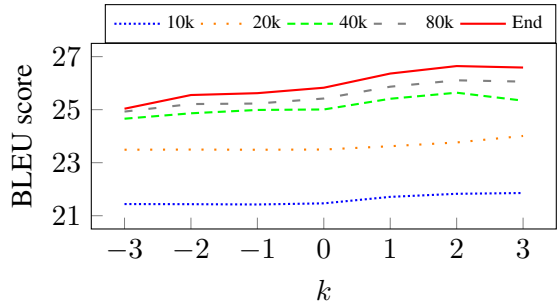
\begin{figure}
    \centering
    \begin{tikzpicture}
    \pgfplotsset{set layers}
    
    \begin{axis}[
        height=0.525 * \columnwidth,
        width=\columnwidth,
        title={},
        xlabel={$k$},
        ytick pos=left,
        xtick pos=bottom,
        scaled x ticks=true,
        xtick scale label code/.code={},
        ylabel={BLEU score},
        xmin=-3.5, xmax=3.5,
        ymin=20.5, ymax=27.5,
        xtick={-4, -3, -2, -1, 0, 1, 2, 3, 4},
        ytick={21, 23, 25, 27},
        grid style=dashed,
        legend cell align=left,
        legend style={
            at={(1.0, 1.0)},
            anchor=south east,
            font=\scriptsize,
			legend columns=5},
		every axis plot/.append style={thick}
    ]
    
    \addplot[
        color=blue,
        densely dotted
        ]
        coordinates {
            (-3, 21.4401)
            (-2, 21.4353)
            (-1, 21.4241)
            (0, 21.4665)
            (1, 21.7112)
            (2, 21.8267)
            (3, 21.8564)
            
        };
    \addlegendentry{10k}
    
    \addplot[
        color=orange,
        loosely dotted
        ]
        coordinates {
            (-3, 23.4892)
            (-2, 23.4953)
            (-1, 23.4882)
            (0, 23.497)
            (1, 23.6212)
            (2, 23.7633)
            (3, 24.0084)
            
        };
    \addlegendentry{20k}
    
    \addplot[
        color=green,
        densely dashed
        ]
        coordinates {
            (-3, 24.6591)
            (-2, 24.8631)
            (-1, 24.9893)
            (0, 25.0073)
            (1, 25.4112)
            (2, 25.64)
            (3, 25.3404)
        };
    \addlegendentry{40k}
    
    \addplot[
        color=gray,
        loosely dashed
        ]
        coordinates {
            (-3, 24.9281)
            (-2, 25.2114)
            (-1, 25.2311)
            (0, 25.4206)
            (1, 25.8661)
            (2, 26.1071)
            (3, 26.0582)
        };
    \addlegendentry{80k}
    
    \addplot[
        color=red
        ]
        coordinates {
            (-3, 25.0357)
            (-2, 25.5523)
            (-1, 25.6221)
            (0, 25.8275)
            (1, 26.3621)
            (2, 26.6444)
            (3, 26.5889)
            
        };
    \addlegendentry{End}
    
    \end{axis}
    
    \end{tikzpicture}
    
    \caption{Affects of $k$ on best performance after certain training steps upon En$\Rightarrow$De dev set.  At the early stage of the training, a higher $k$ yields better translation quality,  denoting a faster convergence speed.}
    \label{fig.searching_k}
\end{figure}

As mentioned in Sec.~\ref{sec.con}, we assign $k$ to balance the extent of discrimination between confident and unconfident examples.
We first conduct experiments on En$\Rightarrow$De to evaluate the impact of $k$. 
Fig.~\ref{fig.searching_k} shows that, the larger $k$, the faster convergence speed.
However, the final performance slightly decreases when $k>2$.
We believe that the overlarge $k$ leads to overfit on confident samples and ignore initial hesitated samples.
This demonstrates that an appropriate balance on the discriminative manner contributes to both convergence acceleration and final performance.

Besides, when $k$ is negative, models will pay more attention to unconfident examples.
This circumstance is identical to reverse-CL \cite{zhang2019curriculum}, where training is advised to offer examples in a hard-to-easy order. 
Our results confirm that unconfidence-first strategy ($k<0$) performs worse than baseline, which is similar with previous findings on CL \cite{zhang2018empirical}.
We attribute this to the fact that the heuristic design forces NMT model to unceasingly learn more from unconfident examples, and finally leads to the strait of catastrophic forgetting~\cite{goodfellow2013empirical}.
Therefore, we set $k=2$ for subsequent experiments.

\subsection{Main Results} 

\begin{table*}[t]
    \centering
    \begin{tabular}{l|lll|c}
        \hline
        \bf Model & \bf{IWSLT15 En$\Rightarrow$Vi} & \bf{WMT14 En$\Rightarrow$De} & \bf{WMT17 Zh$\Rightarrow$En} & \bf{Acc.} \\
        \hline
        \hline
        \textsc{Transformer}-\textit{Base} & ~~~~~30.05 $\pm$ 0.14 & ~~~~27.90 $\pm$ 0.24 & ~~~~24.11 $\pm$ 0.10 & - \\
        ~~~+\textsc{CL-sl} & ~~~~~29.91 $\pm$ 0.13 & ~~~~27.99 $\pm$ 0.22 & ~~~~24.10 $\pm$ 0.08 & 1.02 \\
        ~~~+\textsc{CL-wr} & ~~~~~30.05 $\pm$ 0.17 & ~~~~28.02 $\pm$ 0.24 & ~~~~24.25 $\pm$ 0.09 & 1.06 \\
        \cdashline{1-5}
        SPL & \bf ~~~~~31.21 $\pm$ 0.15$\uparrow$ & \bf ~~~~28.87 $\pm$ 0.19$\uparrow$ & \bf ~~~~24.86 $\pm$ 0.12$\uparrow$ & 1.46 \\
        ~~~w/o TLC & ~~~~~30.91 $\pm$ 0.17$\uparrow$ & ~~~~28.51 $\pm$ 0.21$\uparrow$ & ~~~~24.62 $\pm$ 0.12$\uparrow$ & 1.17 \\
        ~~~w/o SLC & ~~~~~31.14 $\pm$ 0.14$\uparrow$ & ~~~~28.73 $\pm$ 0.24$\uparrow$ & ~~~~24.79 $\pm$ 0.10$\uparrow$ & 1.28 \\
        \hline
        \hline
        \textsc{Transformer}-\textit{Big} & ~~~~~30.61 $\pm$ 0.12 & ~~~~28.72 $\pm$ 0.23 & ~~~~24.57 $\pm$ 0.14 & - \\
        \cdashline{1-5}
        SPL & \bf ~~~~~31.45 $\pm$ 0.15$\uparrow$ & \bf ~~~~29.68 $\pm$ 0.25$\uparrow$ & \bf ~~~~25.26 $\pm$ 0.15$\uparrow$ & 1.21 \\
        \hline
    \end{tabular}
    
    \caption{Overall experimental results of all approaches upon three translation tasks. Each cell contains the mean value and standard variance of BLEU scores derived from 5 independent experimental runs. ``SPL'': proposed self-paced learning model. ``Acc.'': Acceleration ratio of training time required to achieve the best performance of baseline. ``$\uparrow$'': the improvement is significant by contrast to \textsc{Transformer}-\textit{Base/Big} baseline model ($p < 0.01$).}
    \label{tab.all_experiments}
\end{table*}

As shown in Tab.~\ref{tab.all_experiments}, our baseline models outperform the reported results~\cite{vaswani2017attention,platanios2019competence} on the same data, making the evaluation convincing.
The proposed self-paced learning method (SPL) achieves better results than existing CL approaches that artificially determine the difficulty (SL or WR), demonstrating the effectiveness of our method.
Specifically, removing either SLC or TLC decreases the translation quality, indicating that two confidence estimations are complementary to each other.
TLC outperforms its SLC counterpart, which confirms that more fine-grained information benefits to the training.
Moreover, our method consistently improves the translation quality with around 1 BLEU score across all involved tasks and multiple model settings.
This shows the universality and effectiveness of SPL on different scales of training data and model sizes. 

\section{Analysis}
In this section, we further investigate how the proposed method exactly affects the NMT model training by conducting experimental analyses on 1) convergence speed, 2) self-paced adjustment and 3) sequential bucketing.
\subsection{Convergence Speed}
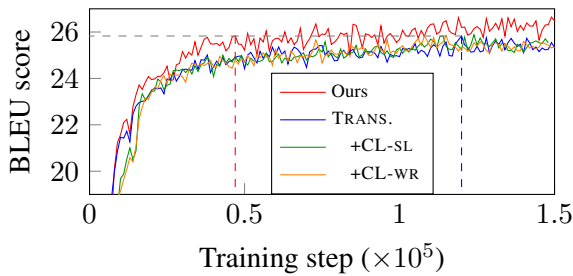
\begin{figure}
    \centering
    \begin{tikzpicture}
    \pgfplotsset{set layers}
    
    \begin{axis}[
        height=0.525 * \columnwidth,
        width=\columnwidth,
        title={},
        xlabel={Training step ($\times 10^{5}$)},
        ytick pos=left,
        xtick pos=bottom,
        scaled x ticks=true,
        xtick scale label code/.code={},
        ylabel={BLEU score},
        xmin=0, xmax=150000,
        ymin=19, ymax=27,
        xtick={0, 50000, 100000, 150000},
        ytick={20, 22, 24, 26},
        grid style=dashed,
        legend cell align=left,
        legend style={
            at={(0.74, 0.0)},
            anchor=south east,
            font=\scriptsize,
			legend columns=1}
    ]
    
    \addplot[
        color=red,
        ]
        coordinates {
            (1000, 0.4715)(2000, 1.0322)(3000, 3.7205)(4000, 9.5983)(5000, 14.6762)(6000, 17.5021)(7000, 17.5838)(8000, 20.071)(9000, 21.2305)(10000, 21.5739)(11000, 22.0967)(12000, 22.2373)(13000, 21.614)(14000, 22.7775)(15000, 23.3011)(16000, 23.7188)(17000, 23.5286)(18000, 23.8334)(19000, 24.0781)(20000, 23.9485)(21000, 23.979)(22000, 24.087)(23000, 24.092)(24000, 24.2463)(25000, 24.5221)(26000, 23.9113)(27000, 23.9378)(28000, 24.5838)(29000, 24.6486)(30000, 25.1556)(31000, 24.8786)(32000, 25.1255)(33000, 25.0041)(34000, 25.5685)(35000, 25.0005)(36000, 25.3564)(37000, 25.6914)(38000, 25.771)(39000, 25.4947)(40000, 25.4767)(41000, 25.5087)(42000, 25.4121)(43000, 25.5668)(44000, 25.1911)(45000, 25.3398)(46000, 25.7766)(47000, 25.8329)(48000, 25.4828)(49000, 25.461)(50000, 25.2171)(51000, 25.9379)(52000, 25.4589)(53000, 25.7097)(54000, 25.0121)(55000, 25.7342)(56000, 25.5715)(57000, 26.044)(58000, 25.1722)(59000, 25.6192)(60000, 25.7214)(61000, 25.727)(62000, 25.5738)(63000, 26.0085)(64000, 25.5349)(65000, 25.7451)(66000, 25.4601)(67000, 25.6099)(68000, 26.1562)(69000, 25.2525)(70000, 25.5147)(71000, 25.709)(72000, 26.0124)(73000, 26.3453)(74000, 25.8009)(75000, 25.9122)(76000, 25.8929)(77000, 26.1409)(78000, 25.4657)(79000, 25.5591)(80000, 25.7119)(81000, 25.9581)(82000, 25.8915)(83000, 25.9546)(84000, 25.0223)(85000, 26.3312)(86000, 26.2875)(87000, 25.8342)(88000, 25.3136)(89000, 25.9208)(90000, 25.8841)(91000, 25.7654)(92000, 25.876)(93000, 25.4452)(94000, 26.2562)(95000, 25.5218)(96000, 25.8182)(97000, 25.935)(98000, 25.7654)(99000, 25.5502)(100000, 25.6801)(101000, 26.0765)(102000, 25.8352)(103000, 25.8135)(104000, 26.0731)(105000, 26.1686)(106000, 25.7406)(107000, 26.0588)(108000, 25.8186)(109000, 26.1703)(110000, 26.327)(111000, 26.3515)(112000, 25.9089)(113000, 26.0067)(114000, 26.4103)(115000, 26.4886)(116000, 26.0496)(117000, 25.9421)(118000, 25.8938)(119000, 26.6145)(120000, 26.1978)(121000, 26.2124)(122000, 25.8671)(123000, 25.9298)(124000, 26.1433)(125000, 26.2329)(126000, 26.1236)(127000, 25.7706)(128000, 26.5771)(129000, 26.0354)(130000, 26.1648)(131000, 26.3769)(132000, 26.1118)(133000, 26.5668)(134000, 25.8269)(135000, 26.606)(136000, 26.2326)(137000, 26.5529)(138000, 26.0237)(139000, 26.2783)(140000, 26.0764)(141000, 26.391)(142000, 26.2591)(143000, 26.2532)(144000, 26.4806)(145000, 25.6492)(146000, 26.0863)(147000, 26.2992)(148000, 26.3215)(149000, 26.6444)(150000, 26.4807)
        };
    \addlegendentry{Ours}
    
    \addplot[
        color=blue,
        ]
        coordinates {
            (1000, 0.4023)(2000, 2.5118)(3000, 4.9833)(4000, 10.9558)(5000, 12.6363)(6000, 17.8103)(7000, 18.7406)(8000, 19.9513)(9000, 20.8653)(10000, 21.4665)(11000, 21.4232)(12000, 21.7248)(13000, 21.2758)(14000, 22.4746)(15000, 22.8074)(16000, 22.9655)(17000, 23.0146)(18000, 23.0349)(19000, 23.2551)(20000, 23.497)(21000, 23.5764)(22000, 23.5501)(23000, 23.5896)(24000, 23.3399)(25000, 23.9732)(26000, 23.9359)(27000, 24.2477)(28000, 23.8822)(29000, 24.0926)(30000, 24.3776)(31000, 24.3995)(32000, 24.3738)(33000, 24.3502)(34000, 24.7658)(35000, 24.4042)(36000, 24.805)(37000, 24.7856)(38000, 25.0073)(39000, 24.2511)(40000, 24.5087)(41000, 24.7462)(42000, 24.5649)(43000, 24.9053)(44000, 24.5925)(45000, 24.7296)(46000, 24.7794)(47000, 24.903)(48000, 24.9554)(49000, 24.9639)(50000, 24.4207)(51000, 25.0071)(52000, 25.1234)(53000, 24.8971)(54000, 24.6711)(55000, 25.1627)(56000, 24.8679)(57000, 25.0826)(58000, 24.9271)(59000, 25.3416)(60000, 25.0273)(61000, 25.3548)(62000, 25.0448)(63000, 25.1796)(64000, 24.6181)(65000, 25.1267)(66000, 25.3496)(67000, 25.269)(68000, 25.1017)(69000, 24.8355)(70000, 24.4756)(71000, 25.1411)(72000, 25.2444)(73000, 25.2663)(74000, 25.4055)(75000, 25.4206)(76000, 25.0152)(77000, 25.1807)(78000, 25.0193)(79000, 24.7628)(80000, 24.7731)(81000, 25.0779)(82000, 25.2115)(83000, 25.0171)(84000, 25.1688)(85000, 25.2578)(86000, 24.9897)(87000, 24.8793)(88000, 24.9121)(89000, 25.1716)(90000, 25.1469)(91000, 25.1467)(92000, 25.0899)(93000, 24.8949)(94000, 25.3711)(95000, 25.0958)(96000, 24.7027)(97000, 25.0398)(98000, 24.9628)(99000, 24.9786)(100000, 24.9402)(101000, 25.295)(102000, 25.2261)(103000, 25.2891)(104000, 25.2379)(105000, 25.4964)(106000, 25.0786)(107000, 25.213)(108000, 25.1415)(109000, 24.9468)(110000, 25.2994)(111000, 25.4517)(112000, 25.3417)(113000, 24.9604)(114000, 25.3973)(115000, 25.5657)(116000, 24.9846)(117000, 24.9648)(118000, 25.4809)(119000, 25.6598)(120000, 25.8275)(121000, 25.4431)(122000, 25.3287)(123000, 25.1358)(124000, 25.1493)(125000, 25.3798)(126000, 25.1444)(127000, 25.1877)(128000, 25.7505)(129000, 25.0943)(130000, 25.4208)(131000, 25.6157)(132000, 25.4649)(133000, 25.6752)(134000, 25.2341)(135000, 25.5472)(136000, 25.7241)(137000, 25.5202)(138000, 25.3785)(139000, 25.4751)(140000, 25.1084)(141000, 25.7277)(142000, 25.5249)(143000, 25.2401)(144000, 25.3539)(145000, 25.1537)(146000, 25.3117)(147000, 25.2294)(148000, 25.2593)(149000, 25.5451)(150000, 25.3344)
        };
    \addlegendentry{\textsc{Trans.}}
    
    \addplot[
        color=black!40!green
        ]
        coordinates {
            (1000, 0.658)(2000, 4.3202)(3000, 8.4603)(4000, 10.3219)(5000, 11.4678)(6000, 16.1236)(7000, 17.4199)(8000, 18.6342)(9000, 18.3833)(10000, 19.6837)(11000, 19.8741)(12000, 20.515)(13000, 21.002)(14000, 20.4977)(15000, 20.8875)(16000, 22.7931)(17000, 23.3371)(18000, 23.076)(19000, 23.261)(20000, 23.4419)(21000, 23.4595)(22000, 23.9652)(23000, 23.8614)(24000, 23.5218)(25000, 23.7099)(26000, 23.7612)(27000, 24.334)(28000, 24.3623)(29000, 24.339)(30000, 24.4083)(31000, 24.2494)(32000, 24.4185)(33000, 24.1897)(34000, 24.5371)(35000, 24.3013)(36000, 24.5951)(37000, 24.6765)(38000, 24.9212)(39000, 24.714)(40000, 24.0823)(41000, 24.6092)(42000, 24.5268)(43000, 24.9969)(44000, 24.6763)(45000, 24.858)(46000, 24.7004)(47000, 24.9512)(48000, 24.7342)(49000, 24.9899)(50000, 24.7152)(51000, 24.5741)(52000, 24.873)(53000, 24.7487)(54000, 24.7288)(55000, 24.738)(56000, 24.8033)(57000, 25.0802)(58000, 24.7203)(59000, 24.7327)(60000, 24.9)(61000, 25.0503)(62000, 24.6705)(63000, 24.8667)(64000, 24.8433)(65000, 24.755)(66000, 24.9195)(67000, 24.9313)(68000, 24.9954)(69000, 25.1052)(70000, 24.7878)(71000, 24.9701)(72000, 24.9343)(73000, 25.0067)(74000, 25.2696)(75000, 25.1373)(76000, 25.5739)(77000, 24.927)(78000, 25.092)(79000, 24.9781)(80000, 24.8776)(81000, 25.2688)(82000, 24.8725)(83000, 24.9729)(84000, 25.0877)(85000, 25.1269)(86000, 25.4214)(87000, 24.9252)(88000, 25.0179)(89000, 25.2763)(90000, 25.0604)(91000, 25.2064)(92000, 25.3561)(93000, 25.3955)(94000, 25.1131)(95000, 25.03)(96000, 25.1901)(97000, 24.9626)(98000, 24.9022)(99000, 25.2345)(100000, 25.3085)(101000, 25.0826)(102000, 25.2927)(103000, 25.1349)(104000, 25.3091)(105000, 25.2134)(106000, 25.3912)(107000, 25.5284)(108000, 25.3525)(109000, 25.4595)(110000, 25.6915)(111000, 25.5987)(112000, 25.3376)(113000, 25.802)(114000, 25.8307)(115000, 25.382)(116000, 25.2603)(117000, 25.2333)(118000, 25.5432)(119000, 25.3931)(120000, 25.2266)(121000, 25.7748)(122000, 25.1929)(123000, 25.4421)(124000, 25.8796)(125000, 25.4485)(126000, 25.7349)(127000, 25.5043)(128000, 25.4977)(129000, 25.3744)(130000, 25.5256)(131000, 25.7081)(132000, 25.3189)(133000, 25.5047)(134000, 25.5001)(135000, 25.1936)(136000, 25.21)(137000, 25.1197)(138000, 25.4751)(139000, 25.3623)(140000, 25.1327)(141000, 25.3258)(142000, 25.5113)(143000, 25.5523)(144000, 25.6177)(145000, 25.5355)(146000, 25.5012)(147000, 25.7059)(148000, 25.5291)
        };
    \addlegendentry{~~~~+\textsc{CL-sl}}
    
    \addplot[
        color=orange
        ]
        coordinates{
            (1000, 0.7016)(2000, 4.4547)(3000, 8.0151)(4000, 9.9913)(5000, 11.9655)(6000, 16.1131)(7000, 17.5983)(8000, 18.2063)(9000, 18.7942)(10000, 19.1525)(11000, 19.6309)(12000, 20.0833)(13000, 20.5803)(14000, 20.4039)(15000, 21.1101)(16000, 22.2598)(17000, 22.7646)(18000, 23.1351)(19000, 23.0961)(20000, 23.418)(21000, 23.624)(22000, 23.5664)(23000, 23.9101)(24000, 24.1107)(25000, 23.7913)(26000, 24.1858)(27000, 24.1306)(28000, 24.145)(29000, 24.142)(30000, 24.0705)(31000, 23.8397)(32000, 24.3546)(33000, 24.6616)(34000, 24.578)(35000, 24.5588)(36000, 24.6102)(37000, 24.5528)(38000, 24.6448)(39000, 24.7644)(40000, 24.5201)(41000, 24.4706)(42000, 24.9464)(43000, 24.7787)(44000, 24.6622)(45000, 24.5046)(46000, 24.4466)(47000, 24.7899)(48000, 24.6199)(49000, 24.8547)(50000, 24.7086)(51000, 24.9876)(52000, 24.7961)(53000, 24.554)(54000, 25.0479)(55000, 24.8967)(56000, 25.1052)(57000, 24.9217)(58000, 24.9174)(59000, 24.5602)(60000, 25.2373)(61000, 24.8721)(62000, 25.0445)(63000, 25.0848)(64000, 24.7475)(65000, 25.1873)(66000, 25.0223)(67000, 24.6889)(68000, 24.9759)(69000, 24.9252)(70000, 25.011)(71000, 24.8992)(72000, 25.2028)(73000, 25.6496)(74000, 25.3083)(75000, 24.6704)(76000, 24.9906)(77000, 25.0497)(78000, 25.2279)(79000, 25.0548)(80000, 25.0458)(81000, 25.1022)(82000, 25.1706)(83000, 25.0096)(84000, 25.3987)(85000, 25.3581)(86000, 25.2582)(87000, 25.2592)(88000, 25.3611)(89000, 25.2089)(90000, 24.9969)(91000, 24.9985)(92000, 25.215)(93000, 25.4634)(94000, 25.249)(95000, 25.1133)(96000, 25.6429)(97000, 25.0978)(98000, 25.1817)(99000, 25.093)(100000, 25.2995)(101000, 25.41)(102000, 25.1898)(103000, 25.1757)(104000, 25.3655)(105000, 25.5634)(106000, 25.7078)(107000, 25.4245)(108000, 25.8259)(109000, 25.6082)(110000, 25.3329)(111000, 25.7601)(112000, 25.3213)(113000, 25.4742)(114000, 25.567)(115000, 25.8434)(116000, 25.475)(117000, 25.4346)(118000, 25.3508)(119000, 25.172)(120000, 25.3558)(121000, 25.3083)(122000, 25.1788)(123000, 25.2412)(124000, 25.5412)(125000, 25.3851)(126000, 25.5818)(127000, 25.4698)(128000, 25.2106)(129000, 25.1045)(130000, 25.4955)(131000, 25.5239)(132000, 25.4628)(133000, 25.1801)(134000, 25.4811)(135000, 25.3836)(136000, 25.2257)(137000, 25.3652)(138000, 25.1896)(139000, 25.1942)(140000, 25.5858)(141000, 25.4675)(142000, 25.3051)(143000, 25.3843)(144000, 25.2701)(145000, 25.3474)(146000, 25.5587)(147000, 25.4875)(148000, 25.4321)(149000, 25.3142)(150000, 25.351)
        };
    \addlegendentry{~~~~+\textsc{CL-wr}}
    
    \addplot[
        color=red,
        dashed
        ]
        coordinates {
            (47000, 25.8329)(47000, 0.00)
        };
        
    \addplot[
        color=blue,
        dashed]
        coordinates {
            (120000, 25.8275)(120000, 0.00)
        };
        
    \addplot[
        color=gray,
        dashed
        ]
        coordinates {
            (-1000, 25.8275)(120000, 25.8275)
        };

    \end{axis}
    
    \end{tikzpicture}
    
    \caption{Learning curves of models on validation set. Our model achieves reductions in iterations of 2.43x.}
    \label{fig.convergence}
\end{figure}

As aforementioned, one motivation of exploiting self-paced learning is to accelerate the convergence of model training.
We visualize the learning curve of examined models on En$\Rightarrow$De dev set in Fig.~\ref{fig.convergence}.
As seen, the vanilla NMT model reaches its convergence at step 120k, while the proposed one gets the same performance at step 47k, yielding 2.43 times faster.
Although Monte Carlo Dropout sampling requires extra time to forward-pass the neural network at each iteration step, our method can still reach comparable result on dev set with shorter training time, achieving 1.46x faster training speed (column ``Acc.'' in Tab.~\ref{tab.all_experiments}).
Besides, we also observe that two methods proposed by~\newcite{platanios2019competence} reveal a comparable tendency with baseline.
We explain this with the view that~\newcite{platanios2019competence} examined these approaches with a batch of 5,120 tokens, much smaller than that used in our experiments (32,768).
Since larger batch size can considerably facilitate the training~\cite{popel2018training,ott2018scaling}, the benefits of their models may be marginal with this change.

\subsection{Self-Paced Adjustment}

\begin{figure}
    \centering
    \begin{tikzpicture}
    \pgfplotsset{set layers}
    
    \begin{axis}[
        height=0.525 * \columnwidth,
        width=0.975 * \columnwidth,
        title={},
        xlabel={Training step ($\times 10^{5}$)},
        ytick pos=left,
        xtick pos=bottom,
        scaled x ticks=true,
        xtick scale label code/.code={},
        ylabel={Ratio},
        xmin=0, xmax=150000,
        ymin=0.94, ymax=1.06,
        xtick={0, 50000, 100000, 150000},
        ytick={0.95, 1.0, 1.05},
        grid style=dashed,
        legend cell align=left,
        legend style={
            at={(1.0, 1.0)},
            anchor=north east,
            font=\scriptsize,
			legend columns=3}
    ]
    
    \addplot[
        color=blue,
        ]
        coordinates {
            (0, 1.0)
            (5000, 1.006932979)
            (10000, 1.002793765)
            (15000, 1.000938578)
            (20000, 1.002774619)
            (25000, 1.002646994)
            (30000, 1.00955438)
            (35000, 0.987201567)
            (40000, 1.003800666)
            (45000, 1.001596492)
            (50000, 0.995567691)
            (55000, 0.997378555)
            (60000, 1.025082551)
            (65000, 0.988643449)
            (70000, 0.963195569)
            (75000, 0.972725738)
            (80000, 0.976565648)
            (85000, 1.010659669)
            (90000, 0.990912069)
            (95000, 1.00169452)
            (100000, 0.990040779)
            (105000, 0.973392812)
            (110000, 0.989317769)
            (115000, 1.003625325)
            (120000, 0.993112813)
            (125000, 0.976438092)
            (130000, 0.995945588)
            (135000, 1.003123697)
            (140000, 0.981389414)
            (145000, 0.98706166)
            (150000, 0.995359186)
        };
    \addlegendentry{short}

    \addplot[
        color=black!40!green,
        ]
        coordinates{
            (0, 1.0)
            (5000, 0.997771497)
            (10000, 0.987936664)
            (15000, 0.992410979)
            (20000, 0.993750107)
            (25000, 1.001348172)
            (30000, 1.003969203)
            (35000, 1.011779133)
            (40000, 0.992112749)
            (45000, 1.003513565)
            (50000, 1.008032046)
            (55000, 1.011553169)
            (60000, 0.979082314)
            (65000, 1.00490438)
            (70000, 1.017386206)
            (75000, 1.005151953)
            (80000, 1.005638522)
            (85000, 0.98575448)
            (90000, 0.990946934)
            (95000, 0.988001696)
            (100000, 0.994214097)
            (105000, 1.027305617)
            (110000, 1.015708662)
            (115000, 1.000542025)
            (120000, 1.002557772)
            (125000, 1.006814466)
            (130000, 0.997505807)
            (135000, 0.98616436)
            (140000, 1.013797921)
            (145000, 1.001378452)
            (150000, 1.003168073)
        };
    \addlegendentry{medium}
    
    \addplot[
        color=red,
        ]
        coordinates{
            (0, 1.0)
            (5000, 0.995332859)
            (10000, 1.009390636)
            (15000, 1.006702186)
            (20000, 1.003504856)
            (25000, 0.996017191)
            (30000, 0.986619948)
            (35000, 1.001171425)
            (40000, 1.004133578)
            (45000, 0.99491037)
            (50000, 0.996448522)
            (55000, 0.991177098)
            (60000, 0.996373007)
            (65000, 1.006550498)
            (70000, 1.020468678)
            (75000, 1.022769744)
            (80000, 1.018255244)
            (85000, 1.003740172)
            (90000, 1.018390838)
            (95000, 1.010431814)
            (100000, 1.01593752)
            (105000, 1.000028097)
            (110000, 0.995164865)
            (115000, 0.996183639)
            (120000, 1.00436601)
            (125000, 1.017198801)
            (130000, 1.006581527)
            (135000, 1.010872093)
            (140000, 1.005095283)
            (145000, 1.011713337)
            (150000, 1.001489659)
            
        };
    \addlegendentry{long}
    
    \addplot[
        color=gray,
        dashed
        ]
        coordinates {
            (0, 1.0)(150000, 1.00)
        };
        
    

    \end{axis}
    
    \end{tikzpicture}
    
    \caption{The ratios between averaged SLC scores gained by our model and baseline. Obviously, the model confidence on training samples with different lengths change constantly during model training.}
    \label{fig.length_scores}
\end{figure}
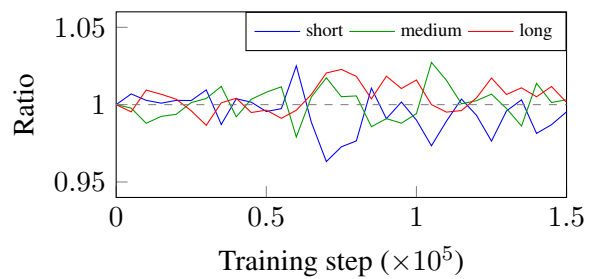
It is interesting to investigate how our model adjusts its learning.
We randomly extract 300 En$\Rightarrow$De training examples, which then be categorized into 3 subsets according to their sentence lengths.
Fig.~\ref{fig.length_scores} shows the ratios of averaged SLC scores between our method and vanilla NMT system at different checkpoints.
As seen, at the beginning of the training, the ratio of confidence score with respect to short sentences is greater than 1, indicating our model pays more attention to shorter examples than baseline.
This is identical with human intuition that the short sentences seem easier and should be learned earlier~\cite{zhang2019curriculum,zhao2020reinforced}.
However, as training continues, our model focuses on short and long sentences simultaneously and hesitates on sentences with medium length, which goes against human intuition and indicates that long sentences may easier than its medium counterparts for current model.
From then on, the curves fluctuate and interlace continuously,  revealing that 
SPL automatically regulates its learning emphasis.
These phenomena show the flexibility of our model, and confirm that predefined data difficulty and learning schedule is insufficient to fully match the model learning.

\subsection{Sequential Bucketing}

Conventional model training sorts examples with similar lengths into buckets to keep efficiency.
This may introduce bias when estimating confidence scores, because longer sequence may gain far less attention due to the productive multiplication of probabilities for SLC estimation.
Generally, larger window size for bucketing increases the diversity of length within each batch, but reduce the efficiency of training due to extra padding tokens. 

\begin{figure}
    \centering
    \begin{tikzpicture}
    \pgfplotsset{set layers}
    
    \begin{axis}[
        height=0.5 * \columnwidth,
        width=\columnwidth,
        title={},
        xlabel={\# of buckets},
        ytick pos=left,
        xtick pos=bottom,
        scaled x ticks=true,
        xtick scale label code/.code={},
        ylabel={BLEU score},
        xmin=0, xmax=80,
        ymin=25, ymax=27,
        xtick={5, 10, 20, 40, 72},
        ytick={25, 26, 27},
        grid style=dashed,
        legend cell align=left,
        legend style={
            at={(0.95, 0.05)},
            anchor=south east,
            font=\scriptsize,
			legend columns=2}
    ]
    
    \addplot[
        color=red,
        mark=x,
        ]
        coordinates {
            (5, 26.6439)
            (10, 26.6549)
            (20, 26.6277)
            (40, 26.6147)
            (72, 26.6943)
            
        };
    \addlegendentry{Ours}
    
    \addplot[
        color=blue,
        mark=+
        ]
        coordinates {
            (5, 25.8511)
            (10, 25.8396)
            (20, 25.8275)
            (40, 25.7292)
            (72, 25.6248)
        };
    \addlegendentry{Baseline}
    
    \end{axis}
    
    \end{tikzpicture}
    
    \caption{Performance upon WMT14 En$\Rightarrow$De dev set with different bucketing strategy. With window size for sequence bucketing being smaller, the number of buckets accordingly increases, and our model can maintain its performance whereas baseline drops.}
    \label{fig.mantissa_bits}
\end{figure}
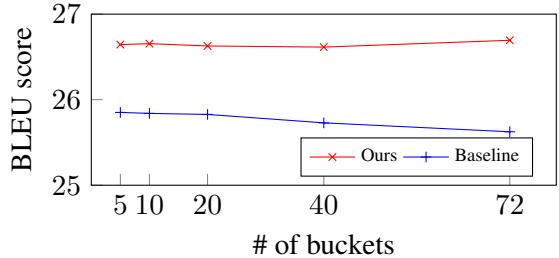

To investigate whether the diversity of sequential lengths within each batch may introduce bias to SLC score computation, we conduct a series of experiments with different settings of sequential bucketing.
As shown in Fig.\ref{fig.mantissa_bits}, we explore the effect of this on En$\Rightarrow$De task, revealing that both baseline and our approach can gain improvement from larger bucket range.
Nevertheless, the performance of baseline model decreases along with lower diversity of sequential lengths, whereas that of our model does not diminish.
Our model gives better performance with smaller window size compared to baseline. Here we can conclude, that the performance of \textsc{Transformer} baseline model is bothered by close sequence lengths within each batch, whereas our model shows its flexibility of adjusting its learning to avoid such effect.

For fair comparison as well as keeping the training efficiency, we follow the default setting from~\newcite{vaswani2017attention} by determining 20 as the number of buckets across all experiments.

\section{Conclusion}
In this paper, we propose a novel self-paced learning model for NMT in which the learning schedule is determined by model itself rather than being intuitively predefined by humans.
Experimental results on three translation tasks verify the universal effectiveness of our approach.
Quantitative analyses confirm that exploiting self-paced strategy presents a more flexible way to facilitate the model convergence than its CL counterparts.
It is interesting to combine with other techniques~\cite{Li:2018:EMNLP,Hao:2019:NAACL} to further improve NMT.
Besides, as this idea is not limited to machine translation, it is also interesting to validate our model in other NLP tasks, such as low-resource NMT model training~\cite{lample2018phrase,wan2020unsupervised} and neural architecture search~\cite{guo2020breaking}.

\section*{Acknowledgement}

This work was supported by National Key R\&D Program of China (2018YFB1403202), the National Natural Science Foundation of China (Grant No.~61672555), the Joint Project of the Science and Technology Development Fund, Macau SAR and National Natural Science Foundation of China (Grant No.~045/2017/AFJ), the Science and Technology Development Fund, Macau SAR (Grant No.~0101/2019/A2), the Multi-year Research Grant from the University of Macau (Grant No. MYRG2020-00054-FST).
This work was performed in part at the Super Intelligent Computing Center supported by State Key Laboratory of Internet of Things for Smart City and the High-Performance Computing Cluster supported by Information and Communication Technology Office of the University of Macau.
We thank the anonymous reviewers for their insightful comments.

\bibliography{emnlp2020}
\bibliographystyle{acl_natbib}

\end{document}